\documentclass[conference]{IEEEtran}
\IEEEoverridecommandlockouts
\usepackage{cite}
\usepackage{amsmath,amssymb,amsfonts}
\usepackage{algorithmic}
\usepackage{graphicx}
\usepackage{textcomp}
\usepackage{makecell}
\usepackage{booktabs}
\usepackage{fancyhdr}
\usepackage{epsfig}
\usepackage{times}
\usepackage{mathtools}
\usepackage{url}
\usepackage{enumitem}
\usepackage{adjustbox}
\usepackage{subcaption}
\usepackage{multirow}
\usepackage[table,xcdraw]{xcolor}
\usepackage{comment}
\usepackage{tikz}
\usepackage{array}

\usepackage[pagebackref,breaklinks,colorlinks]{hyperref}

\def\BibTeX{{\rm B\kern-.05em{\sc i\kern-.025em b}\kern-.08em
    T\kern-.1667em\lower.7ex\hbox{E}\kern-.125emX}}
\begin{document}

\title{MorDIFF: Recognition Vulnerability and Attack Detectability of Face Morphing Attacks Created by Diffusion Autoencoders\\
\thanks{This research work has been funded by the German Federal Ministry of Education and Research and the Hessian Ministry of Higher Education, Research, Science, and the Arts within their joint support of the National Research Center for Applied Cybersecurity ATHENE.}
}

\author{\parbox{16cm}{\centering
    {\large Naser Damer$^{1,2}$},  Meiling Fang$^{1,2}$, Patrick Siebke$^{1}$, Jan Niklas Kolf$^{1,2}$ \\ Marco Huber$^{1,2}$, Fadi Boutros$^{1}$  \\
    {\normalsize
    $^1$Fraunhofer Institute for Computer Graphics Research IGD, Darmstadt, Germany\\
   $^2$Department of Computer Science, TU Darmstadt, Darmstadt, Germany \\
    }}
}
\maketitle

\begin{abstract}
Investigating new methods of creating face morphing attacks is essential to foresee novel attacks and help mitigate them.
Creating morphing attacks is commonly either performed on the image-level or on the representation-level.
The representation-level morphing has been performed so far based on generative adversarial networks (GAN) where the encoded images are interpolated in the latent space to produce a morphed image based on the interpolated vector.
Such a process was constrained by the limited reconstruction fidelity of GAN architectures.
Recent advances in the diffusion autoencoder models have overcome the GAN limitations, leading to high reconstruction fidelity.
This theoretically makes them a perfect candidate to perform representation-level face morphing. 
This work investigates using diffusion autoencoders to create face morphing attacks by comparing them to a wide range of image-level and representation-level morphs.
Our vulnerability analyses on four state-of-the-art face recognition models have shown that such models are highly vulnerable to the created attacks, the MorDIFF, especially when compared to existing representation-level morphs.
Detailed detectability analyses are also performed on the MorDIFF, showing that they are as challenging to detect as other morphing attacks created on the image- or representation-level. Data and morphing script are made public \footnote{https://github.com/naserdamer/MorDIFF}.
\end{abstract}


\vspace{-2mm}
\section{Introduction}
\label{sec:intro}
\vspace{-1mm}
%The deep-learning driven performance advances in face recognition \cite{DBLP:conf/cvpr/DengGXZ19}, along with the relatively high social acceptance \cite{Jain:1998:BPI:552539}, have brought automatic face recognition to be a key technology in security sensitive applications of identity management (e.g. travel documents) \cite{FaceMarket2017}. 

Face recognition (FR) systems, despite their high accuracy, are vulnerable to many attacks, one of which is face morphing attack.
Face morphing attacks aim at creating face images that are verifiable to be the face of multiple identities, which can lead to building faulty identity links in operations like border checks.
This was shown by  Ferrara et al. \cite{DBLP:conf/icb/FerraraFM14} when they proved that one attack image can successfully match more than one person, automatically and by human experts.
When such morphing attacks are used in association with travel or identity documents, it can allow multiple subjects to verify their identity to the alphanumerical information of the document. 
This possible false link to the identity information might enable a number of illegal actions related to financial transactions, human trafficking, illegal immigration, among others.

Face morphing can be performed on the image-level, commonly by interpolating facial landmarks in the morphed images and blending the texture information \cite{DBLP:conf/icb/RaghavendraRVB17,DBLP:journals/tifs/FerraraFM18}.
Morphing can also be performed on the representation-level by interpolating face image representations and decoding this interpolated representation into a face morphing attack.
This representation-level morphing has been so far performed using GAN architectures, with latent (representation) size depending on the number of GAN training samples, leading to low reconstruction fidelity.
Typically, despite being prone to blending artifacts, the image-level morphing led to attacks that much more strongly preserve the identities of the morphed images, in comparison to GAN-based representation-level morphs.
With recent advancements in the diffusion autoencoders \cite{DBLP:conf/nips/HoJA20,DBLP:conf/icml/Sohl-DicksteinW15,DBLP:conf/cvpr/PreechakulCWS22}  have aimed at avoiding the pitfalls of GAN architectures to significantly enhance the reconstruction fidelity.
This makes diffusion autoencoders a perfect candidate to perform representation-level morphing with exceptional identity-preservation and realistic image appearance.

This work investigates the vulnerability of state-of-the-art FR models to morphing attacks created on the representation-level by diffusion autoencoders.
To perform this, we created our diffusion-based morphing attacks, the MorDIFF, by interpolating the semantics and stochastic latent representation of the two face images to be morphed. 
This interpolated latent is later decoded into the morphing attack image.
We investigated the vulnerability of a set of FR models to MorDIFF attacks, proving their high ability to attack FR systems in comparison to other morphing techniques.
We additionally investigated the detectability of MorDIFF attacks by a set of morphing attack detection (MAD) solutions based on different backbones and training data, pointing out the challenging nature of detecting MorDIFF attacks, among other attacks.

\vspace{-2mm}
\section{Related works}
\label{sec:rw}
\vspace{-1mm}
Face morphing is either performed on the image-level or on the representation level. 
First face morphing attacks were created on the image-level by detecting facial landmarks in the source images to be morphed. 
The facial landmarks are interpolated and the texture is blended, resulting in what is commonly referred to as landmark-based morphs (LMA). 
Variations of this process were used in the literature, such as the work of Ferrera et al. \cite{DBLP:journals/tifs/FerraraFM18} and Ramachandra et al. \cite{DBLP:conf/icb/RaghavendraRVB17}. 
An early comparison \cite{DBLP:journals/iet-bmt/ScherhagKRB20} of such methods has shown that the approach used in  \cite{DBLP:conf/icb/RaghavendraRVB17,DBLP:conf/dagm/DamerBWBTBK18} achieved the relatively strong face morphing attacks, i.e. high identity preservation of the morphed identities. Further variations of this process performed the interpolation individually on partial facial parts \cite{DBLP:journals/tbbis/QinPVRLB21}, producing attacks that are harder to detect by the MAD. 
The listed image-level morphs, and others that are used in the experiment of this work, have various degrees of image artifacts introduced by the fact that the identity interpolation is performed on the image-level \cite{DBLP:journals/tbbis/ZhangVRRDB21}. 

Given the advances in GAN architectures and their performance in producing synthetic images, and to avoid the disadvantages of image-level interpolation, MorGAN GAN-based morphing approach was proposed in \cite{DBLP:conf/btas/DamerS0K18}. The first representation-level face morphing attacks were MorGAN, where the solution encoded the images to be morphed into the GAN latent space and interpolated these latent representations. The interpolated latent vector is then decoded by the GAN generator to produce the morphed face image. These early representation-level morphs preserved the identities moderately and were of low resolution, but were proven to be hard to detect if they were not used to train the MAD \cite{DBLP:conf/fusion/DamerZWSKK19,DBLP:conf/btas/DamerGZKK19}. A post-generation cascaded enhancement step was added in a follow-up work on the MorGAN generator to increase the image perceptual quality, nonetheless, with the same identity preservation level \cite{DBLP:conf/btas/DamerBSKK19}.
Based on the concept introduced by MorGAN \cite{DBLP:conf/btas/DamerS0K18}, Venkatesh et al. created morphed images with better identity preservation qualities and a much more realistic appearance  \cite{DBLP:conf/iwbf/VenkateshZRRDB20}. This was mainly driven by the use use of the StyleGAN architecture by Karras et al. \cite{DBLP:conf/cvpr/KarrasLA19}. Later on, also based on the StyleGAN architecture \cite{DBLP:conf/cvpr/KarrasLA19}, the MIPGAN I and II were introduced to generate images with higher identity preservation \cite{DBLP:journals/tbbis/ZhangVRRDB21}. This was enforced by a training loss function that optimizes the identity interpolation on the latent-level. A hybrid family of morphing approaches did use the GAN architecture, but as a post-processing step after performing the morphing on the image-level. ReGenMorphs \cite{DBLP:conf/isvc/DamerRSVBFKRK21} introduced this concept to take advantage of the identity-preservation qualities of the image-level morphing while encoding and decoding the morphed image in a GAN architecture to produce a new face (without blending artifacts) that posses the morphed identity information. 

The existing image-level morphing techniques produce the desired identity-preservation properties of a strong attack, however, are linked to visible blending artifacts.
The existing representation-level morphing techniques are all GAN-based. Such a method preserves the desired identity to a much lower degree than image-based morphs and is linked to synthetic-like generation artifacts when the manipulations are performed on the latent space \cite{DBLP:conf/btas/DamerS0K18,DBLP:conf/iwbf/VenkateshZRRDB20,DBLP:journals/tbbis/ZhangVRRDB21}.
This is caused by the relatively low reconstruction fidelity of GAN architectures, which are themselves dependent on the size of training data \cite{DBLP:conf/cvpr/PreechakulCWS22}.
This is what diffusion models targeted by achieving higher fidelity and thus act as a perfect candidate to facilitate more realistic and more identity-preserving representation-level morphing attacks, as investigated in this work. 
A recent not-peer-reviewed pre-print was posted a few days before the submission of this work where such a concept was investigated \cite{DBLP:journals/corr/abs-2301-04218}, however with vulnerabilities being measured with extremely outdated and outperformed FR solutions, making the presented results practically irrelevant.
Regardless of the morphing techniques, all possible morphing attacks should be considered and novel attacks should be foreseen to help build defensive mechanisms before such attacks are used by actual attackers, especially for highly realistic and strong attacks.

%Although the existing LMA morphs have strong identity preservation capabilities, the fact that they build their identity blend on the image level makes them prone to image artifacts. The GAN-based morphs do generally preserve the identity to a lower degree \cite{DBLP:journals/tbbis/ZhangVRRDB21}, however, this is less relevant in such a scenario where the worst-case attack scenario needs to be considered. Despite the enhanced quality of the GAN-based morphed images, the manipulation in the latent space still produce synthetic-like generation artifacts \cite{DBLP:conf/btas/DamerS0K18,DBLP:conf/iwbf/VenkateshZRRDB20,DBLP:journals/tbbis/ZhangVRRDB21}. This work aims to eliminate the LMA blending artifacts by using a GAN-based generation, as well as, eliminate the manipulation in the latent space, resulting in visibly realistic morphed images compared to previous works.

%end need re-write

\vspace{-2mm}
\section{Diffusion-based face morphing}
\label{sec:morph}
\vspace{-1mm}
This section describes the morphing technique used to create the MorDIFF morphs. 
The approach is based on the work of Preechakul et al. \cite{DBLP:conf/cvpr/PreechakulCWS22} and is described in the following with the necessary background information.

Previous morphing generation process used the GAN architecture \cite{DBLP:conf/btas/DamerS0K18,DBLP:conf/btas/DamerBSKK19,DBLP:conf/iwbf/VenkateshZRRDB20} and generated good morph attacks. Despite the fact that GAN can learn meaningful representations in the latent spaces, the reconstruction fidelity is limited due to the latent size depending on the number of training samples. In contrast to GAN, diffusion probabilistic model (DPM) \cite{DBLP:conf/nips/HoJA20,DBLP:conf/icml/Sohl-DicksteinW15} has achieved higher visual fidelity in image generation. However, the latent code of DPM lacks semantic meaning, which makes it harder for diverse representation learning in comparison with GAN-based models. To address such issues, diffusion autoencoder \cite{DBLP:conf/cvpr/PreechakulCWS22} was proposed to separately infer both semantic and stochastic information from the inputs. Therefore, we use the diffusion autoencoder \cite{DBLP:conf/cvpr/PreechakulCWS22} to generate high fidelity and smooth face morphing between two input face images $x_I^1$ and $x_I^2$.

The diffusion autoencoder contains two encoders. One is a semantic encoder mapping $x_I^1$ and $x_I^2$ to semantic latent representations $\mathbf{z}_s^1$ and $\mathbf{z}_s^2$. The other one is a stochastic encoder to obtain the latent code $\mathbf{x}_t^1$ and $\mathbf{x}_t^2$ from inputs. Therefore, two subcodes are encoded from $x_I^1$ and $x_I^2$ to $(\mathbf{z}_s^1, \mathbf{x}_t^1)$ and $(\mathbf{z}_s^2, \mathbf{x}_t^2)$. Then, a linear interpolation $Lerp$ and a spherical linear interpolation $SLerp$ (as recomendded in \cite{DBLP:conf/siggraph/Shoemake85} and \cite{DBLP:conf/iclr/SongME21}) are used to obtain a representative latent code $\mathbf{z(\lambda)}$ with $\lambda \in [0,1]$ from the obatined subcodes. The interpolation can be formulated as following:
\begin{equation}
    Lerp(\mathbf{z}_s^1, \mathbf{z}_s^2; \lambda) = \lambda \mathbf{z}_s^1 + (1-\lambda) \mathbf{z}_s^2
\end{equation}
\begin{equation}
    SLerp(\mathbf{x}_t^1, \mathbf{x}_t^2; \lambda) = \frac{\sin \theta \lambda \mathbf{x}_t^1 }{\sin \theta} + \frac{\sin \theta (1-\lambda) \mathbf{x}_t^2}{\sin \theta} + 
\end{equation}
where 
\begin{equation}
\theta =  \frac{ \arccos( \mathbf{x}_t^1 \cdot \mathbf{x}_t^2 ) }{ \|\mathbf{x}_t^1\| \|\mathbf{x}_t^2\|}
\end{equation}.

\begin{figure*}[h!]
    \centering
    \includegraphics[width=0.81\textwidth]{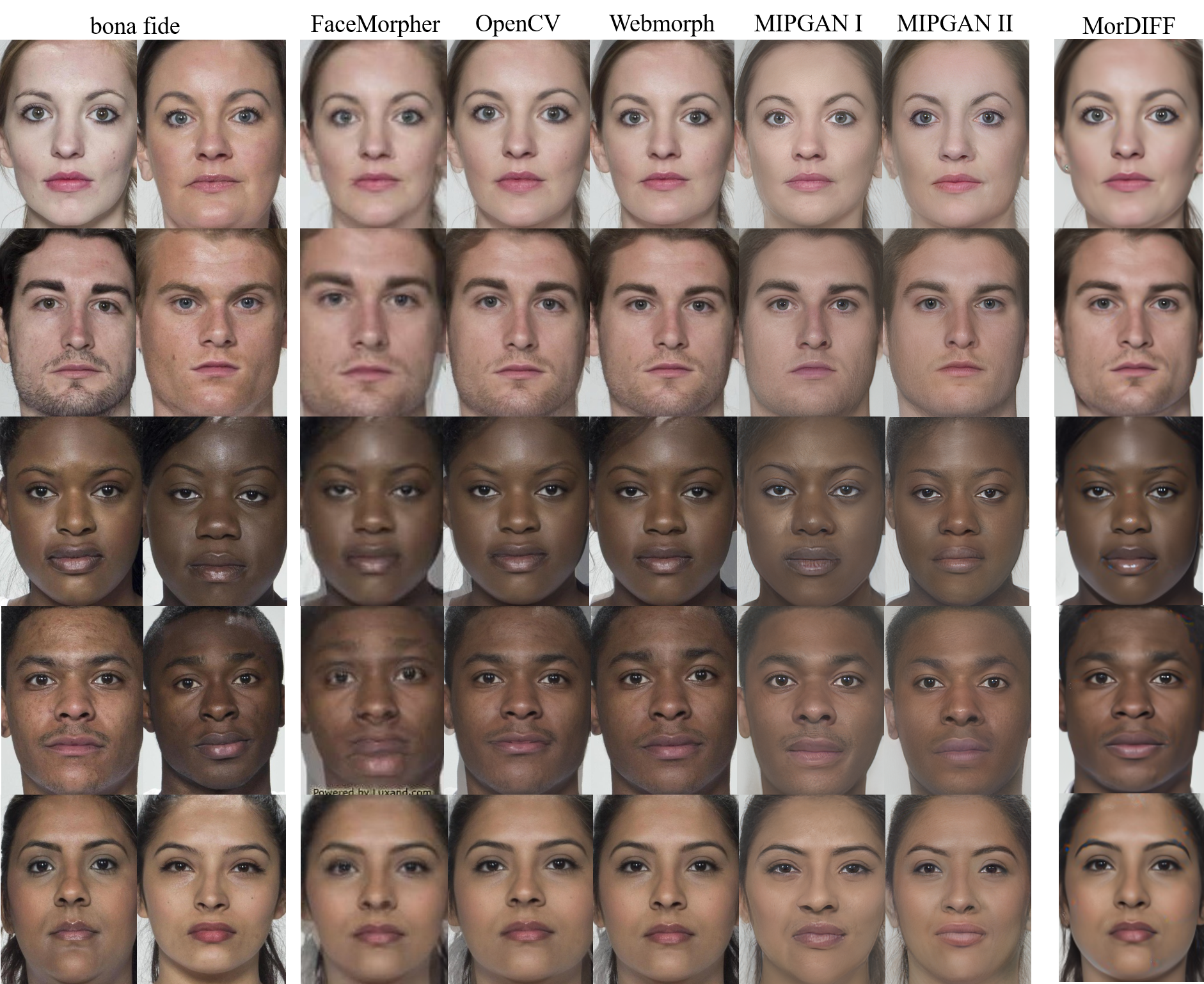}
    \caption{Samples of the MorDIFF (right most column) attacks, the baseline attacks (created by FaceMorpher, OpenCV, WebMorph, MIPGANI and II), and the bona fide images that were morphed to creat the attack (two left most columns). The image-level morphs (FaceMorpher, OpenCV, WebMorph) show the traditional blending artifacts, while the representation-level morphs (MIPGAN-I and II) show typical streaking GAN artifacts. The MorDIFF attacks in these samples show much lower level of generative artifacts, as expected when comparing diffusion-based generation to GAN-based generation \cite{DBLP:conf/cvpr/PreechakulCWS22}. }
    \label{fig:sample}
    \vspace{-3mm}
\end{figure*}

Therefore, the latent code $\mathbf{z(\lambda)}$ can formulated as $\mathbf{z(\lambda)} = (Lerp(\mathbf{z}_s^1, \mathbf{z}_s^2; \lambda), SLerp(\mathbf{x}_t^1, \mathbf{x}_t^2; \lambda))$. Finally, a conditional DPM is used as decoder to produce the morphing attack, denoted as MorDIFF in our case.
The used diffusion autoencoder training details are provided in \cite{DBLP:conf/cvpr/PreechakulCWS22} and we used the pre-trained autoencoder available publicly \footnote{https://github.com/phizaz/diffae}.

In the whole morphing process, semantic encoder is able to manipulate the facial features for partial identity information preserving and stochastic encoder provides the detailed information to produce high visual fidelity morphing attack samples, as shown in Figure \ref{fig:sample}.

\vspace{-2mm}
\section{The MorDIFF dataset} %DONE % DONE Grammerly
\label{sec:exp:db}
\vspace{-1mm}
The MorDIFF dataset extends over the SYN-MAD 2022 competition dataset \cite{DBLP:conf/icb/HuberBLRRDNGSCT22} and uses the same morphing pairs to enable a comparable dataset set. The MorDIFF and the SYN-MAD 2022 datasets are both based on the Face Research Lab London (FRLL) dataset \cite{DeBruine2021}. The FRLL dataset contains images of 102 different individuals and provides high-quality frontal images created in a controlled scenario with a wide range of different ethnicities. All individuals present in the dataset signed consent for their images to be "used in lab-based and web-based studies in their original or altered forms and to illustrate research." For the morph generation, we limited the data to the frontal images of the dataset, similar to SYN-MAD 2022.

The pairs are selected by splitting the frontal images of the FRLL dataset depending on the provided gender and expression (neutral or smiling). ElasticFace-Arc \cite{DBLP:conf/cvpr/BoutrosDKK22} is then used to generate embeddings of the images and these embeddings are then compared with cosine similarity within the split to find the most similar faces. The 250 most similar face pairs are selected, resulting in 250 female neutral pairs, 250 female smiling pairs, 250 male neutral pairs, and 250 male smiling pairs, in total 1000 MorDIFF attack images and the 204 bona fide images of the SYN-MAD 2022 dataset.

The SYN-MAD 2022 dataset, which we use as a benchmark, contains morphed images from 5 different approaches, three image-level (FaceMorpher (commercial-of-the-shelf), OpenCV \cite{openCVmorph} and Webmorph (online tool \footnote{https://webmorph.org/} ) and two representation-level GAN-based (MIPGAN I \cite{DBLP:journals/tbbis/ZhangVRRDB21} and MIPGAN II \cite{DBLP:journals/tbbis/ZhangVRRDB21}). We follow the same morph pair selection protocol (described above) and now add the MorDIFF attacks to the benchmark, which will be publicly released.
Other GAN-based representation-level morphing techniques, such as the MorGAN \cite{DBLP:conf/btas/DamerS0K18}, EMorGAN \cite{DBLP:conf/btas/DamerBSKK19}, and StyleGAN morphs \cite{DBLP:conf/iwbf/VenkateshZRRDB20} were not considered, as their quality and identity preservation ability is far inferior MIPGAN I and II as shown in \cite{DBLP:journals/tbbis/ZhangVRRDB21}).

Samples of the MorDIFF attacks are shown in Figure \ref{fig:sample} along with the morphed bonafide images and the morphing attacks created by the five baseline methods. 
Looking carefully at the images, one notices that the image-level morphs (OpenCV, FaceMorpherm, and WebMorph) do, as expected have higher shadowing and blending artifacts in comparison to the representation-level morphs (MIPGAN I and II, and the MorDIFF). 
The MIPGAN I and II contain streaking artifacts common to the GAN-generated faces. This kind of artifact is almost non-existent in the samples of the MorDIFF attacks.

We considered measuring the statistical and perceptual image quality of the different morphs and found that previous works have shown no clear correlation between the image quality and the realistic appearance when dealing with Morphing attacks (typically of ICAO standard \cite{ICAO} and of very high perceived quality) \cite{DBLP:journals/tbbis/ZhangVRRDB21}. Such investigation have proven to lead to faulty conclusions, and thus we opted out of such study. 
In \cite{DBLP:journals/tbbis/ZhangVRRDB21}, the PSNR and SSIM values showed an insignificant difference between the different attacks, and a minimum difference was inconsistent with the perceived visual quality. Other work showed that the visibly unrealistic images of MorGAN have higher statistical quality metrics (6 different metrics  \cite{DBLP:conf/btas/DamerBSKK19}) than other attacks.  In a recent study, researchers have shown that operations that apparent morphing artifacts do not consistently affect the estimated quality across a large number of quality estimation strategies \cite{DBLP:conf/biosig/FuMorph21,https://doi.org/10.1049/bme2.12094}, however, especially utility metrics, can be used to differentiate between attacks and bona fide to a small degree \cite{https://doi.org/10.1049/bme2.12094}. However, studies on the human observer's ability to detect the MorDIFF attack are necessary and are planned to evaluate the relative perceived quality of these attacks, as conducted on previous attack methods in \cite{9997091}.

\vspace{-2mm}
\section{Vulnerability of face recognition to MorDIFF attacks}
\label{sec:vuln}
\vspace{-1mm}
We evaluated the vulnerability of four FR systems to the MorDIFF attacks in comparison to the five baseline attacks. The chosen FR systems are two of the top-performing large-scale FR solution, i.e. ElasticFace (ElasticFace-Arc) \cite{DBLP:conf/cvpr/BoutrosDKK22} and CurricularFace \cite{DBLP:conf/cvpr/HuangWT0SLLH20}, and two top performing light-weight FR solutions that can be implemented in embedded and portable devices, i.e. the MixFaceNet \cite{DBLP:conf/icb/BoutrosDFKK21} and PocketNet \cite{DBLP:journals/access/BoutrosSKDKK22}. CurricularFace and ElasticFace are based on ResNet-100 architectures and have $55.52$M parameters each with $24192.51$ MFLOPs, MixFaceNet consists of $3.07$M parameters with $451.7$ MFLOPs and PocketNet has $1.75$M parameters and $1099.15$ MFLOPs. On the frequently used benchmark Labeled Faces in the Wild \cite{CPLFWTech}, CurricularFace and ElasticFace achieve both an accuracy of $99.80$\%, MixFaceNet scores $99.60$\% and PocketNet $99.58$\%. On the AgeDB-30 benchmark \cite{moschoglou2017agedb}, the accuracy is respectively $98.32$\% for CurricularFace, $98.35$\% for ElasticFace, $96.63$\% for MixFaceNet and $97.17$\% for PocketNet. All the FR models used are the official releases by the respective authors.

We present the vulnerability results in a quantifiable manner by listing the Mated
Morphed Presentation Match Rate (MMPMR) \cite{DBLP:conf/biosig/ScherhagNRGVSSM17} and the Fully Mated Morphed Presentation Match Rate (FMMPMR) \cite{DBLP:conf/biosig/ScherhagNRGVSSM17} based on the decision threshold at the false match rate of (FMR) 1\% and 0.1\%, MMPMR100 and MMPMR1000, respectively. 

% \usepackage{multirow}
\begin{table*}[]
\centering
\resizebox{0.99\textwidth}{!}{
\begin{tabular}{|cl|ll|ll|ll|ll|}
\hline
\multicolumn{2}{|c|}{FR model }                                                & \multicolumn{2}{c|}{ElasticFace \cite{DBLP:conf/cvpr/BoutrosDKK22}}                               & \multicolumn{2}{c|}{CurricularFace \cite{DBLP:conf/cvpr/HuangWT0SLLH20}}                            & \multicolumn{2}{c|}{MixFaceNet \cite{DBLP:conf/icb/BoutrosDFKK21}}                                & \multicolumn{2}{c|}{PocketNet \cite{DBLP:journals/access/BoutrosSKDKK22}}                                 \\ \hline
\multicolumn{2}{|c|}{Morphing technique }                                      & \multicolumn{1}{c|}{MMPMR100} & \multicolumn{1}{c|}{MMPMR1000} & \multicolumn{1}{c|}{MMPMR100} & \multicolumn{1}{c|}{MMPMR1000} & \multicolumn{1}{c|}{MMPMR100} & \multicolumn{1}{c|}{MMPMR1000} & \multicolumn{1}{c|}{MMPMR100} & \multicolumn{1}{c|}{MMPMR1000} \\ \hline
\multicolumn{1}{|c|}{\multirow{3}{*}{Image level}}          & OpenCV          & \multicolumn{1}{l|}{\textit{0.997}}    & 0.980                          & \multicolumn{1}{l|}{\textit{0.996}}    & 0.986                          & \multicolumn{1}{l|}{\textit{0.996}}    & 0.963                          & \multicolumn{1}{l|}{\textit{0.996}}    & 0.970                          \\ \cline{2-10} 
\multicolumn{1}{|c|}{}                                      & FaceMorpher     & \multicolumn{1}{l|}{0.962}    & 0.913                          & \multicolumn{1}{l|}{0.970}    & 0.935                          & \multicolumn{1}{l|}{0.972}    & 0.931                          & \multicolumn{1}{l|}{0.979}    & 0.941                          \\ \cline{2-10} 
\multicolumn{1}{|c|}{}                                      & WebMorph        & \multicolumn{1}{l|}{0.990}    & \textit{0.986}                          & \multicolumn{1}{l|}{0.988}    & \textit{0.988}                          & \multicolumn{1}{l|}{0.988}    & \textit{0.984}                          & \multicolumn{1}{l|}{0.988}    & \textit{0.988}                          \\ \hline \hline
\multicolumn{1}{|c|}{\multirow{3}{*}{Representation level}} & MIPGAN-I        & \multicolumn{1}{l|}{0.980}    & 0.845                          & \multicolumn{1}{l|}{0.962}    & 0.890                          & \multicolumn{1}{l|}{0.981}    & 0.887                          & \multicolumn{1}{l|}{0.991}    & 0.900                          \\ \cline{2-10} 
\multicolumn{1}{|c|}{}                                      & MIPGAN-II       & \multicolumn{1}{l|}{0.953}    & 0.778                          & \multicolumn{1}{l|}{0.953}    & 0.832                          & \multicolumn{1}{l|}{0.973}    & 0.836                          & \multicolumn{1}{l|}{0.977}    & 0.857                          \\ \cline{2-10} 
\multicolumn{1}{|c|}{}                                      & MorDIFF (our) & \multicolumn{1}{l|}{\textbf{0.990}}    & \textbf{0.948}                          & \multicolumn{1}{l|}{\textbf{0.995}}    & \textbf{0.968}                          & \multicolumn{1}{l|}{\textbf{0.992}}    & \textbf{0.958}                          & \multicolumn{1}{l|}{\textbf{0.996}}    & \textbf{0.949}                          \\ \hline
\end{tabular}
}
\vspace{-1mm}
\caption{Vulnerability of four face recognition solutions to the MorDIFF and the baseline attacks. Under all evaluation settings, the MorDIFF attacks were stronger (FR more vulnerable) than the other representation-level morphing attacks (indicated by the highest MMPMR values in this category in bold). The vulnerability to the MorDIFF attacks is, in many evaluation settings, very close to the strongest image-level morphing attack (highest MMPMR values indicated by italic font in that category).}
\vspace{-4mm}
\label{tab:vuln}
\end{table*}

Table \ref{tab:vuln} presents the results of our vulnerability analyses.
On all four considered FR solutions, and for both operating points (FMR of 1\% and 0.1\%), the FR is more vulnerable to MorDIFF attacks in comparison to other representation-level attacks.
This is evident by the large difference between the MMPMR values (MMPMR100 and MMPMR1000) scored by the MorDIFF attacks in comparison to those scored by the MIPGAN I and II attacks.
As expected, FR solutions are more vulnerable to the blending artifact-prone image-level morphing techniques in comparison to the MIPGAN I and II attacks, especially the image-level morphs created by OpenCV and WebMorph.
This gap between the FR vulnerability to image-level and representation-levels morphs became very tight when comparing the image-level morphs to MorDIFF attacks.
MorDIFF attacks score MMPMR values that are in most cases extremely close to the best-performing image-level morphs. One can also notice that for all attack types, and especially for lower FMR (0.1\% ,i.e. MMPMR1000) the vulnerability of the slightly higher performing large-scale FR models (ElasticFace and CutticularFace) is higher than the vulnerability of the compact light-weight models (MixFaceNet and PocketNet).

\vspace{-2mm}
\section{Detectability of MorDIFF attacks}
\vspace{-1mm}
To measure the detectability of the MorDIFF attacks, along with the baseline attacks, we deploy two well-performing MADs: MixFaceNet-MAD and Inception-MAD. 

MixFaceNet-MAD employs the backbone of MixFaceNet \cite{DBLP:conf/icb/BoutrosDFKK21} for face verification and identification. Considering its lower computation complexity (FLOPs) and high accuracy, MixFaceNet was used successfully for face MAD \cite{SMDD} by adding a classification layer after embedding layer to make the MAD decision. We followed the same modification and settings in \cite{SMDD} to conduct experiments.
Inception-MAD uses the Inception-v3 \cite{inceptionv3} network architecture as a backbone. This architecture has been used successfully for MAD \cite{DBLP:conf/cvip/RamachandraVRB18,DBLP:conf/isvc/DamerSFBKK21}.
Both models are trained from scratch by using the cross-entropy loss function and Adam optimizer with a learning rate of $10^{-4}$ and a weight decay of $10^{-5}$, where the input size for Inception-MAD is $229 \times 229 \times 3$ and for MixFaceNet-MAD is $224 \times 224 \times 3$. 
To avoid overfitting and for a fair comparison, an early stopping technique with the patience of 20 epochs.

Three instances of each of the MADs are evaluated, and each of these instances is trained on a different database. 
This results in a realistic evaluation protocol where all the MADs are trained on datasets of different origins than the evaluation ones (different source datasets and different identities).
The three training datasets used in experiments are SMDD, LMA-DRD, and MorGAN-LMA. 

SMDD \cite{SMDD} dataset is a synthetic-based MAD dataset that 25,000 bona fide images are created by StyleGAN2-ADA \cite{DBLP:conf/nips/KarrasAHLLA20,DBLP:conf/iwbf/VenkateshZRRDB20}, and 15,000 morph attacks are generated based on such bona fide by using OpenCV morphing technique \cite{openCVmorph}. Considering the privacy-friendly characteristic and applicability of SMDD for generalized MADs, we use SMDD as one of the training sets.
In addition to synthetic MAD data, we utilized two real MAD datasets, LMA-DRD and MorGAN-LMA. 
LMA-DRD includes 123 bona fide samples selected from VGGFace2 dataset \cite{DBLP:conf/fgr/CaoSXPZ18} and 88 morphed attacks generated by OpenCV morphing \cite{openCVmorph}. These attacks were then created by the re-digitized (print and scan) method. 
MorGAN-LMA contains 750 bona fides selected from CelebA \cite{DBLP:conf/iccv/LiuLWT15} and 500 morphs created by OpenCV landmark-based morphing \cite{openCVmorph}.

The MAD performance (detectability) is presented by the Attack Presentation Classification Error Rate (APCER), i.e. the proportion of attack images incorrectly classified as bona fide samples, at a fixed (1\%, 10\%, and 20\%) Bona fide Presentation Classification Error Rate (BPCER), i.e. the proportion of bona fide images incorrectly classified as attack samples, as defined in the ISO/IEC 30107-3 \cite{ISO301073}. Additionally, the Detection Equal Error Rate (EER), i.e. the value of APCER or BPCER at the decision threshold where they are equal, is reported. 

% \usepackage{multirow}
\begin{table}[]
\resizebox{0.48\textwidth}{!}{
\centering
\begin{tabular}{|c|c|l|l|lll|}
\hline
\multicolumn{1}{|l|}{\multirow{2}{*}{MAD}} & \multicolumn{1}{l|}{\multirow{2}{*}{Train data}} & \multirow{2}{*}{Test data} & \multirow{2}{*}{EER (\%)} & \multicolumn{3}{c|}{APCER (\%) @ BPCER (\%)}                       \\ \cline{5-7} 
\multicolumn{1}{|l|}{}                     & \multicolumn{1}{l|}{}                            &                            &                           & \multicolumn{1}{l|}{1.00}   & \multicolumn{1}{l|}{10.00}  & 20.00  \\ \hline
\multirow{18}{*}{\rotatebox[origin=c]{90}{MixFaceNet-MAD}}           & \multirow{6}{*}{\rotatebox[origin=c]{90}{SMDD}}                            & FaceMorph                  & 4.60                      & \multicolumn{1}{l|}{5.50}   & \multicolumn{1}{l|}{3.60}   & 2.90   \\ \cline{3-7} 
                                           &                                                  & MIPGAN\_I                  & 16.70                     & \multicolumn{1}{l|}{75.80}  & \multicolumn{1}{l|}{22.20}  & 14.50  \\ \cline{3-7} 
                                           &                                                  & MIPGAN\_II                 & 20.62                     & \multicolumn{1}{l|}{81.58}  & \multicolumn{1}{l|}{32.03}  & 20.62  \\ \cline{3-7} 
                                           &                                                  & OpenCV                     & 8.33                      & \multicolumn{1}{l|}{36.38}  & \multicolumn{1}{l|}{6.50}   & 3.76   \\ \cline{3-7} 
                                           &                                                  & WebMorph                   & 18.20                     & \multicolumn{1}{l|}{74.00}  & \multicolumn{1}{l|}{24.00}  & 17.60  \\ \cline{3-7} 
                                           &                                                  & MorphDiffusion             & 8.50                      & \multicolumn{1}{l|}{33.40}  & \multicolumn{1}{l|}{7.40}   & 4.10   \\ \cline{2-7} 
                                           & \multirow{6}{*}{\rotatebox[origin=c]{90}{LMAD-DRD}}                   & FaceMorph                  & 5.60                      & \multicolumn{1}{l|}{11.20}  & \multicolumn{1}{l|}{3.30}   & 1.20   \\ \cline{3-7} 
                                           &                                                  & MIPGAN\_I                  & 14.40                     & \multicolumn{1}{l|}{64.90}  & \multicolumn{1}{l|}{19.80}  & 7.70   \\ \cline{3-7} 
                                           &                                                  & MIPGAN\_II                 & 11.51                     & \multicolumn{1}{l|}{48.65}  & \multicolumn{1}{l|}{13.51}  & 4.30   \\ \cline{3-7} 
                                           &                                                  & OpenCV                     & 16.37                     & \multicolumn{1}{l|}{72.46}  & \multicolumn{1}{l|}{25.61}  & 12.09  \\ \cline{3-7} 
                                           &                                                  & WebMorph                   & 21.60                     & \multicolumn{1}{l|}{82.80}  & \multicolumn{1}{l|}{46.60}  & 23.80  \\ \cline{3-7} 
                                           &                                                  & MorphDiffusion             & 21.40                     & \multicolumn{1}{l|}{81.60}  & \multicolumn{1}{l|}{43.90}  & 22.30  \\ \cline{2-7} 
                                           & \multirow{6}{*}{\rotatebox[origin=c]{90}{MorGAN-LMA}}                      & FaceMorph                  & 8.00                      & \multicolumn{1}{l|}{13.90}  & \multicolumn{1}{l|}{7.30}   & 0.00   \\ \cline{3-7} 
                                           &                                                  & MIPGAN\_I                  & 14.60                     & \multicolumn{1}{l|}{93.00}  & \multicolumn{1}{l|}{30.40}  & 0.00   \\ \cline{3-7} 
                                           &                                                  & MIPGAN\_II                 & 18.92                     & \multicolumn{1}{l|}{93.19}  & \multicolumn{1}{l|}{34.33}  & 0.00   \\ \cline{3-7} 
                                           &                                                  & OpenCV                     & 9.86                      & \multicolumn{1}{l|}{78.76}  & \multicolumn{1}{l|}{13.41}  & 0.00   \\ \cline{3-7} 
                                           &                                                  & WebMorph                   & 15.80                     & \multicolumn{1}{l|}{92.40}  & \multicolumn{1}{l|}{34.00}  & 0.00   \\ \cline{3-7} 
                                           &                                                  & MorphDiffusion             & 13.50                     & \multicolumn{1}{l|}{89.30}  & \multicolumn{1}{l|}{27.20}  & 0.00   \\ \hline
\multirow{18}{*}{\rotatebox[origin=c]{90}{Inception-MAD}}            & \multirow{6}{*}{\rotatebox[origin=c]{90}{SMDD}}                            & FaceMorph                  & 0.00                      & \multicolumn{1}{l|}{1.70}   & \multicolumn{1}{l|}{0.00}   & 0.00   \\ \cline{3-7} 
                                           &                                                  & MIPGAN\_I                  & 10.90                     & \multicolumn{1}{l|}{50.90}  & \multicolumn{1}{l|}{13.70}  & 5.70   \\ \cline{3-7} 
                                           &                                                  & MIPGAN\_II                 & 16.22                     & \multicolumn{1}{l|}{82.48}  & \multicolumn{1}{l|}{25.83}  & 11.41  \\ \cline{3-7} 
                                           &                                                  & OpenCV                     & 7.52                      & \multicolumn{1}{l|}{28.66}  & \multicolumn{1}{l|}{5.49}   & 3.05   \\ \cline{3-7} 
                                           &                                                  & WebMorph                   & 18.00                     & \multicolumn{1}{l|}{85.20}  & \multicolumn{1}{l|}{27.40}  & 13.40  \\ \cline{3-7} 
                                           &                                                  & MorphDiffusion             & 5.30                      & \multicolumn{1}{l|}{17.20}  & \multicolumn{1}{l|}{3.50}   & 2.50   \\ \cline{2-7} 
                                           & \multirow{6}{*}{\rotatebox[origin=c]{90}{LMAD-DRD}}                   & FaceMorph                  & 61.00                     & \multicolumn{1}{l|}{99.90}  & \multicolumn{1}{l|}{99.60}  & 97.40  \\ \cline{3-7} 
                                           &                                                  & MIPGAN\_I                  & 41.30                     & \multicolumn{1}{l|}{99.60}  & \multicolumn{1}{l|}{89.50}  & 70.40  \\ \cline{3-7} 
                                           &                                                  & MIPGAN\_II                 & 39.74                     & \multicolumn{1}{l|}{99.60}  & \multicolumn{1}{l|}{88.19}  & 63.37  \\ \cline{3-7} 
                                           &                                                  & OpenCV                     & 8.84                      & \multicolumn{1}{l|}{40.96}  & \multicolumn{1}{l|}{8.23}   & 2.13   \\ \cline{3-7} 
                                           &                                                  & WebMorph                   & 20.20                     & \multicolumn{1}{l|}{85.60}  & \multicolumn{1}{l|}{41.00}  & 17.00  \\ \cline{3-7} 
                                           &                                                  & MorphDiffusion             & 95.40                     & \multicolumn{1}{l|}{100.00} & \multicolumn{1}{l|}{100.00} & 100.00 \\ \cline{2-7} 
                                           & \multirow{6}{*}{\rotatebox[origin=c]{90}{MorGAN-LMA}}                      & FaceMorph                  & 0.80                      & \multicolumn{1}{l|}{0.70}   & \multicolumn{1}{l|}{0.00}   & 0.00   \\ \cline{3-7} 
                                           &                                                  & MIPGAN\_I                  & 46.10                     & \multicolumn{1}{l|}{98.70}  & \multicolumn{1}{l|}{89.20}  & 78.00  \\ \cline{3-7} 
                                           &                                                  & MIPGAN\_II                 & 35.84                     & \multicolumn{1}{l|}{96.80}  & \multicolumn{1}{l|}{74.28}  & 56.26  \\ \cline{3-7} 
                                           &                                                  & OpenCV                     & 9.34                      & \multicolumn{1}{l|}{45.43}  & \multicolumn{1}{l|}{9.25}   & 2.95   \\ \cline{3-7} 
                                           &                                                  & WebMorph                   & 19.00                     & \multicolumn{1}{l|}{70.00}  & \multicolumn{1}{l|}{31.80}  & 18.40  \\ \cline{3-7} 
                                           &                                                  & MorphDiffusion             & 26.50                     & \multicolumn{1}{l|}{87.00}  & \multicolumn{1}{l|}{53.10}  & 32.50  \\ \hline
\end{tabular}}
\vspace{-1mm}
\caption{The detectability of the MorDIFF attacks and the baseline attacks using six different MADs (two architectures and 3 training datasets). The MorDIFF attacks are challenging to detect, as are other attack types, especially in the realistic scenario of cross-dataset evaluation we present here. Please note that the rows with EER values above 50\% (which is not realistic) indicate that the detection scores of the MAD, in such cases, indicated that the bona fide are closer to being predicted as attacks than the attacks themselves.}
\vspace{-4mm}
\label{tab:det}
\end{table}

The detectability results are presented in Table \ref{tab:det}. Keeping in mind that all the MAD evaluation scenarios presented here are realistic cross-dataset evaluations, all the morphing attacks were not perfectly detected.
MADs trained on the large-scale synthetic-based SMDD tend to perform better than other MADs, which might be due to the higher diversity in the training data.
For most MADs, the MorDIFF was not the most, nor the least accurately detected morph type.
Thus they can be considered to be challenging, similarly to other attack types, to be detected by the MADs.
In general, the MixFaceNet-MAD performed slightly better than the Inception-MAD in most cases, whether to detect MorDIFF or other attacks, keeping in mind that this MAD (trained on SMDD) was the baseline for the recent SYN-MAD 2022 competition \cite{DBLP:conf/icb/HuberBLRRDNGSCT22}.
MorDiff attacks seem to be also detected relatively better when the MADs are trained on the SMDD dataset.

\vspace{-2mm}
\section{Conclusion}
\label{sec:con}
\vspace{-1mm}
This paper investigated using diffusion autoencoders to generate morphing attacks by interpolating the semantics and stochastic latent representation of the two face images to be morphed, which are later decoded to create a morphed image, i.e. the MorDIFF. 
We showed that state-of-the-art FR solutions are extremely vulnerable to this representation-level morphing technique, especially when compared to the recent GAN-based representation-level morphing solutions such as MIPGAN I and II.
The MorDIFF attacks are also challenging to detect, similarly to other morphing attacks.
The MorDIFF data will be publicly released and it follows the same generation protocol of the recent MAD evaluation benchmark, SYN-MAD 2022, to maintain research reproducibility and comparison.
Future research will look into the generalization benefits of including the MorDIFF morphs in the training of MAD solutions.

\vspace{-2mm}
\bibliographystyle{plain}
\bibliography{strings,refs}

\end{document}